%% file: main.tex
\documentclass{article} 
\usepackage{preprint_style,times}

\input{math_commands.tex}

\usepackage{amssymb}

\usepackage{alltt}
\usepackage[hidelinks]{hyperref}
\usepackage{xurl}
\usepackage{booktabs}
\usepackage{graphicx}
\usepackage{multirow}
\usepackage{xspace}
\usepackage{enumitem}
\usepackage{xcolor}
\usepackage[most]{tcolorbox}
\newtcblisting{promptbox}[1]{listing only, breakable, enhanced, colback=gray!4, colframe=black!55,
  boxrule=0.4pt, arc=1pt, left=3pt, right=3pt, top=2pt, bottom=2pt,
  title={#1}, fonttitle=\small\bfseries, coltitle=white, colbacktitle=black!55,
  listing options={basicstyle=\ttfamily\scriptsize, breaklines=true, breakatwhitespace=false,
    columns=fullflexible, keepspaces=true, showstringspaces=false}}

\newcommand{\absval}[1]{\textcolor{gray}{#1}}

\newif\ifshowcomments
\showcommentsfalse

\newcommand{\ourselect}{DefUse\xspace}

\title{Between Spoilers and Guesswork: Avoiding Over-Specification in Coding Benchmarks}

\author{%
Yuanzhe Liu$^1$\thanks{Work was done while the first author interned at IBM.} \quad
Bastian Lipka$^2$
\quad
Avraham Shinnar$^2$
\quad
Jatin Ganhotra$^2$
\\
\textbf{Yao Ma}$^1$
\quad
\textbf{Martin Hirzel}$^2$
\\
$^1$ Rensselaer Polytechnic Institute \quad
$^2$ IBM
\\
\texttt{\{liuy72,may13\}@rpi.edu}
\\
\texttt{Lipka.Bastian@ibm.com}
\\
\texttt{\{shinnar,jatinganhotra,hirzel\}@us.ibm.com}
}

\preprintfinalcopy 
\begin{document}

\maketitle
\lhead{Between Spoilers and Guesswork: Avoiding Over-Specification in Coding Benchmarks}

\input{sections/abstract}
\input{sections/introduction}
\input{sections/related_work}
\input{sections/methodology}

\input{sections/experiments}

\input{sections/conclusions}
\bibliography{ref}
\bibliographystyle{preprint_style}

\appendix
\input{sections/appendix}

\end{document}

%% file: math_commands.tex
\usepackage{amsmath,amsfonts,bm}

\def\eqref#1{equation~\ref{#1}}

\def\1{\bm{1}}

\newcommand{\test}{\mathcal{D_{\mathrm{test}}}}

\DeclareMathAlphabet{\mathsfit}{\encodingdefault}{\sfdefault}{m}{sl}
\SetMathAlphabet{\mathsfit}{bold}{\encodingdefault}{\sfdefault}{bx}{n}



%% file: sections/abstract.tex
\begin{abstract}
Benchmarks drive AI progress, but are hampered by task
under-specification and over-specification.
Task under-specification causes guesswork, where solutions have to
make things up, or worse, resort to contamination or reward hacking.
Task over-specification causes spoilers, where solutions can bypass
some of the harder steps the benchmark was supposed to measure.
Since future-looking benchmarks should be difficult,
over-specification reduces a benchmark's shelflife of community
adoption.
Repository-level coding benchmarks have helped the field progress but
experience unnecessary churn due to both over- and
under-specification.
For instance, SWE-bench Pro, FEA-bench, Doc2Feat-bench, and Multi-SWE-bench all tackle under-specification by augmenting task
statements with hints derived from ground truth.
We show that leaking ground-truth information into the task
statement causes over-specification in some of these benchmarks.
We experiment with various specification levels, trying to thread the needle between spoilers and guesswork.
Based on these results, we recommend techniques for deriving
parsimonious hints.
We will release our hints so the community can obtain more reliable
signals from popular benchmarks.
\end{abstract}

%% file: sections/introduction.tex
\section{Introduction}








Repository-level coding benchmarks have contributed to great advancements in coding 
agents by evaluating them via executing held-out tests against their proposed patches~\citep{jimenez2024swebench}. 
For these evaluations to measure implementation ability, the task description and available repositories have to communicate the requirements enforced by the tests.
Achieving this alignment has proven difficult.
OpenAI introduced SWE-bench Verified to reduce underspecification and unfair testing~\citep{openai_2024_verified}. 
It later recommended SWE-bench Pro after finding test flaws and contamination in Verified~\citep{openai_2026_pro}, then withdrew that recommendation after finding underspecified prompts and overly restrictive tests in Pro~\citep{openai_2026}.
This churn is evidence for both the importance and difficulty of this problem.
One source of misalignment is a test that requires a particular new function, class, file, or attribute name that the agent cannot infer from the task description or repository.
The problem is especially relevant to feature-addition benchmarks~\citep{li2025fea,deng2025nocode}, where the new names  may be entirely missing in the existing repository.
An agent can then implement the requested behavior correctly yet fail solely because it chose a different name.
Recovering these unstated interface choices introduces \emph{guesswork}
into an evaluation intended to measure software implementation ability. 

\begin{figure}
  \begin{minipage}{0.53\textwidth}
    \centerline{\includegraphics[scale=0.5]{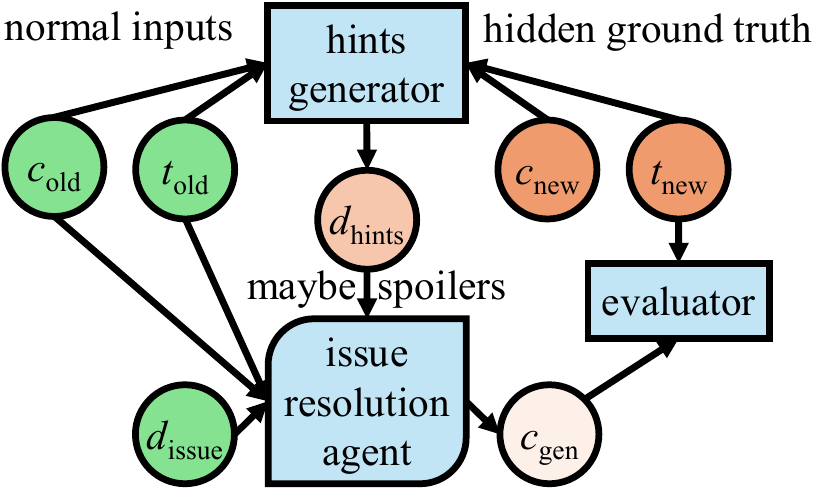}}
    \caption{\label{fig:hints_overview}To hint (perchance spoilers) or not to hint~(guesswork): that is the question for repository-level coding benchmarks.}
  \end{minipage}
  \hspace*{\fill}
  \begin{minipage}{0.45\textwidth}
    \centerline{\includegraphics[scale=0.5]{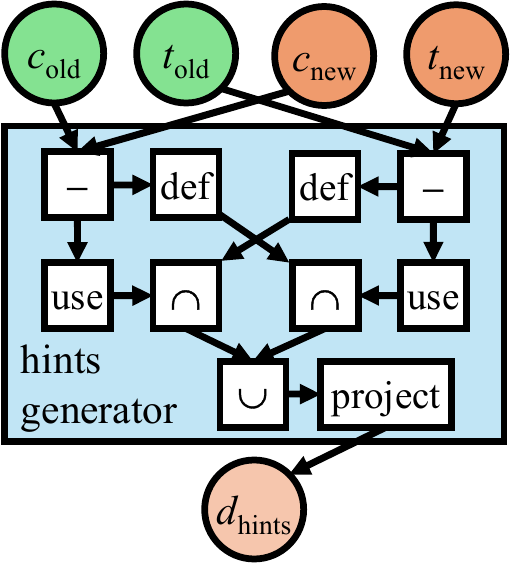}}
    \caption{\label{fig:hints_def_use}DefUse hints generation pipeline.}
  \end{minipage}
\end{figure}

To reduce this guesswork, several benchmarks augment task descriptions with information derived from reference implementations and test patches~\citep{deng2025nocode,li2025fea,deng2025swe,badertdinov_et_al_2025}.
We refer to these additions as \emph{hints}. 
Figure~\ref{fig:hints_overview} provides an overview of how hints are generated and used in repository-level coding benchmarks.
An agent receives a task description $d_{\mathrm{issue}}$ and existing code and tests $(c_{\mathrm{old}},t_{\mathrm{old}})$,
and produces code $c_{\mathrm{gen}}$ evaluated against held-out tests $t_{\mathrm{new}}$.
A hint generator additionally accesses the reference code $c_{\mathrm{new}}$ and held-out tests to produce
$d_{\mathrm{hints}}$, an extra input to the agent.
Hints can expose details that an agent needs to satisfy the evaluation, but the reference implementation also contains design decisions that the benchmark may intend the agent to make independently.
Disclosing these decisions introduces \emph{spoilers}, i.e., information that supplies part of the implementation work being evaluated.
Prior work documents solution leakage through issue reports and comments as a major problem for SWE-bench~\citep{aleithan2024swebenchplus}.
Therefore, identifier hint construction faces a tension between \emph{under-specification}, which leaves agents guessing about hidden constraints, and \emph{over-specification}, which leaks parts of the solution.
Higher solve rates alone cannot distinguish between resolving guesswork and supplying spoilers.
Our goal is to study identifier hint constructions that clarify the interface required by the tests while leaving implementation choices to the agent.

\paragraph{A design space for hints.}
We consider hint construction through two design decisions: \emph{identifier selection} and \emph{identifier projection}.
Selection determines \emph{which} program entities are disclosed, such as file changes, functions edits, etc.
Identifier projection asks \emph{how much} is revealed about each selected entity.
For instance, if a function identifier is selected, then its name, its signature, its location, or its semantics descriptions are all valid information to be considered being revealed. 
Existing hint schemes combine different choices along both dimensions~\citep{deng2025nocode,li2025fea,deng2025swe}, making it difficult to attribute performance differences to either choice when comparing complete schemes.
Separating selection from projection allows us to examine these choices through controlled comparisons.



Guided by this distinction, this paper proposes \textbf{DefUse} for identifier selection and \textbf{Declaration} for identifier projection.
DefUse targets identifiers through which the implementation and its tests interact.
Declaration reveals interface attributes while omitting semantic descriptions and implementation guidance.
Together, these policies aim to reduce \emph{guesswork} about the required identifiers while limiting \emph{spoilers} about how to implement it.

We compare selection and projection policies across five repository-level coding benchmarks spanning Python and Java.
We complement these comparisons with hint attribute ablations and an LLM-based assessment on SWE-bench Pro, examining both task resolution and disclosure of reference-solution information. 
Our findings indicate that performance gains can accompany greater disclosure of implementation choices, motivating evaluation beyond solve rates alone.
Together, these findings motivate combining DefUse selection with Declaration projection to reduce \emph{guesswork} about required interfaces while limit \emph{spoilers} about implementation choices.

The source code of our hint generation pipeline and the resulting hints are available open-source: \url{https://github.com/IBM/swe-hints/}.

\paragraph{Contributions.}
This paper makes three contributions:
\begin{itemize}[topsep=0mm,parsep=0mm]
    \item It characterizes identifier hints through selection and projection, providing a common framework for comparing existing benchmark policies.

    \item It introduces DefUse and Declaration, two policies designed to expose test-facing interfaces while limiting additional reference-derived information.

    \item It reports controlled comparisons and attribute ablations across five benchmarks and three language models, examining how hint construction affects measured task resolution. 
     Additionally, it adopts an LLM judge to assess information necessity, solution disclosure, and behavioral guidance.
\end{itemize}

To the extent that benchmarks drive AI progress, we hope that finding the right balance between spoilers and guesswork in coding benchmarks will ultimately contribute to more reliable coding assistants.

%% file: sections/related_work.tex
\section{Related Work}

\input{tables/hint_schemes_v2}

\paragraph{Benchmark validity.}
Audits of SWE-bench and SWE-bench Pro document underspecified tasks and tests enforcing unstated implementation details~\citep{openai_2024_verified,openai_2026,openai_2026_pro}.
SWE-Bench+ identifies solutions disclosed in issue reports or comments, alongside inadequate tests~\citep{aleithan2024swebenchplus}.
Differential testing and manual analysis further show that passing benchmark tests does not guarantee patch correctness~\citep{wang_pradel_liu_2026}.
Evidence of memorization raises concerns about training-data contamination~\citep{liang_garg_moghaddam_2026}, which LiveCodeBench and SWE-rebench seek to mitigate by collecting new tasks~\citep{jain_et_al_2025,badertdinov_et_al_2025}.
Our study complements these audits by examining how reference-derived hints affect the information available to agents at evaluation time.

\paragraph{Task specification and hints.}
\citet{khatib2026makes} examine how bug-report content affects repair success, including controlled ablations of benchmark-added requirements and interfaces.
\citet{ganhotra_2026} identifies \emph{hidden naming contracts}, where tests require reference-solution identifiers absent from the issue and base repository.
Benchmarks address this ambiguity with reference-derived names, declarations, or requirements~\citep{deng2025nocode,li2025fea,deng2025swe}.
Within agent workflows, SpecRover infers code intent~\citep{ruan_zhang_roychoudhury_2024},
REAgent refines requirements~\citep{kuang_et_al_2026}, and HULA incorporates developer feedback into plans and code~\citep{takerngsaksiri_et_al_2025}.
Interactive approaches address missing requirements through clarification: Ambig-SWE evaluates agents' ability to detect underspecification and elicit missing information~\citep{vijayvargiya2026ambigswe}.
LHAW systematically removes task information and measures its effects through agent trials~\citep{pu2026lhaw}.
Our study separates \emph{identifier selection} from \emph{identifier projection} to 
examine which benchmark disclosures reduce \emph{guesswork} about required interfaces and which supply \emph{spoilers} about implementation choices.


%% file: tables/hint_schemes_v2.tex
\begin{table*}[t]
\centering
\normalsize
\setlength{\tabcolsep}{2pt}
\renewcommand{\arraystretch}{1.05}
\newcommand{\hintcell}[1]{\begin{tabular}[t]{@{}c@{}}#1\end{tabular}}
\caption{What each hint scheme discloses. DefUse is ours improved from Doc2Feat. Doc2Feat means the identifier hints extraction used in the Doc2Feat benchmark. FEA-Bench presents two hint selection: brief and detailed. In SWE Bench Pro, the authors assign "Requirements" as the essential hints, and make "Interface" optional.}
\label{tab:hint-schemes-v2}
\small
\begin{tabular}{@{}p{2.6cm}llccccccc@{}}
\toprule
 & \multicolumn{2}{c}{\textbf{Selection}}
 & \multicolumn{7}{c}{\textbf{Projection}} \\
\cmidrule(lr){2-3}\cmidrule(lr){4-10}
\textbf{Hint scheme} & Author & Rule
 & Name & Type & Path & Signature & Semantics & Behaviour & \shortstack{Document\\Change} \\
\midrule
DefUse
 & auto & def--use
 & \checkmark & \checkmark & \hintcell{new\\files} & \checkmark & --- & --- & --- \\
\addlinespace[2pt]
Doc2Feat
 & auto & \hintcell{new $\cap$\\test}
 & \checkmark & \checkmark & \hintcell{new\\files} & --- & --- & --- & --- \\
\addlinespace[2pt]
FEA-Bench Brief

 & auto & \hintcell{new\\defs}
 & \checkmark & \checkmark & \checkmark & \checkmark & --- & --- & --- \\
\addlinespace[2pt]
FEA-Bench \newline Detailed
 & auto & \hintcell{new\\defs}
 & \checkmark & \checkmark & \checkmark & \checkmark & docstring & --- & \checkmark \\
\addlinespace[2pt]
SWE-bench Pro\newline Interface
 & human & manual
 & \checkmark & \checkmark & \checkmark & \checkmark & \checkmark & --- & --- \\
\addlinespace[2pt]
SWE-bench Pro\newline Interface +\newline Requirements
 & human & manual
 & \checkmark & \checkmark & \checkmark & \checkmark & \checkmark & \checkmark & --- \\
\bottomrule
\end{tabular}
\end{table*}

%% file: sections/methodology.tex
\section{Methodology}
We characterize identifier hints through two decisions: \emph{Identifier Selection}, which determines which program entities are disclosed, and \emph{Identifier Projection}, which
determines which attributes of each selected entity are revealed.
Table~\ref{tab:hint-schemes-v2} organizes existing hint schemes along these dimensions and situates our proposed selection policy, DefUse, within this design space.
This decomposition supports controlled comparisons of selection policies~(under a fixed projection) and projection policies~(under a fixed selection).

\subsection{Identifier Selection: DefUse}
\label{sec:defuse}

Various identifier selection pipelines exist in the current repo-level coding benchmarks. The SWE-Bench Pro authors employ human engineers to construct requirements and interface, while FEA-Bench includes all new components present in the new files between commits~\citep{deng2025swe, li2025fea}. 
The Doc2Feat pipeline first extracts identifiers that appear after new code patches and new test patches, then selects identifiers from the intersection of those extracted entities between code and test patches~\citep{deng2025nocode}. 
Although Doc2Feat has the strictest identifier selection across available pipelines, occurrence in both changes does not distinguish where an identifier is defined from where it is used, leading to including irrelevant information such as field accesses and general imports.

Thus, we propose DefUse, where the selected identifiers are defined and used. Figure~\ref{fig:hints_def_use} demonstrates DefUse pictorially.
If first uses the difference (aka.~the patch) between old and new to find newly introduced names:
\mbox{$c_\textrm{new}-c_\textrm{old}$} and \mbox{$t_\textrm{new}-t_\textrm{old}$}.
Then, DefUse keeps only those names that carry a communication between code and tests, in one of two directions:
\begin{enumerate}[topsep=0mm,parsep=0mm]
    \item \mbox{$\textrm{def}(c_\textrm{new}-c_\textrm{old})\cap\textrm{use}(t_\textrm{new}-t_\textrm{old})$}: The code defines the name and the test uses it; $\cup$
    \item
      \mbox{$\textrm{def}(t_\textrm{new}-t_\textrm{old})\cap\textrm{use}(c_\textrm{new}-c_\textrm{old})$}: The test defines the name and the code uses it.
\end{enumerate}
These are the names through which the solution and its tests communicate. 
The held-out tests have already committed to them, and the agent cannot recover them from the issue alone.

\paragraph{Why defined and used?}
The criterion follows from what a held-out test can actually observe. 
A test rejects a functionally correct patch only through the identifiers it references in that patch: the functions it imports and calls, the classes it instantiates, the attributes and return structures it inspects. 
This yields a necessity argument and a sufficiency argument for DefUse.

\paragraph{Necessity.}
If the test patch reads an identifier that the code patch defines (scenario~1), then the identifier's name is an unforced design choice of the reference author, yet the test treats it as a hard constraint. 
Every such identifier is a potential spurious failure: an agent can implement the required behaviour perfectly and still fail on an exact-match comparison against an arbitrary label. 
These identifiers must therefore be disclosed, or the benchmark scores guesswork rather than capability. 
The symmetric direction (scenario~2) is a contract in the same sense: the test constructs an input the code must consume, and the code-side name is equally unrecoverable and equally load-bearing.

\paragraph{Sufficiency.} 
Conversely, an identifier that is not both defined in one patch and used in the other never crosses the code--test boundary.
The agent may name it, place it, or eliminate it freely without affecting any test outcome. 
Disclosing it repairs no ambiguity, where it potentially leaks the reference author's internal design decisions, which is precisely the spoilers we aim to avoid. \ourselect is thus the minimal selection that removes name-mismatch artifacts. 
Every selected identifier can cause a false failure if hidden, and no discarded identifier can.


The goal of identifier hints is to address ambiguity in the current coding tasks. 
We aim to provide agents as few hints as possible to avoid unnecessary information leakage because exposing too much information will make the problems lose the functionality of evaluating coding agents' true capabilities. 
DefUse follows this principle; it discloses the full interface contract and nothing else.

\subsection{Identifier Projection: Declaration}
\label{sec:declaration}
Declaration reveals, for each selected identifier, what its declaration line states
and nothing more: its name; its kind (function, class, variable, or file); the path of
the file that declares it; whether that file is newly added; and, for callables, its
signature. 
The principle is that a hint should reproduce what a colleague could read off a header or a stub, namely identity, kind, location, and invocation, and stop where the implementation begins. 
For example, for the SWE-bench Pro task that adds locale detection to Ansible, the Declaration entry for the new function reads
\begin{quote}\small\begin{alltt}
### `get_best_parsable_locale` (function)
    -- `lib/ansible/module_utils/common/locale.py` [new file]
Signature: `get_best_parsable_locale(module, preferences=None)`
\end{alltt}\end{quote}
which fixes the name the hidden tests import, where it must live, and how it is called,
and leaves everything else to the agent. 


Declaration deliberately omits the semantic attributes that richer schemes attach to
each entity: a one-line summary of what the entity does and a paragraph describing how.
For the same function, the full hint goes on to name the shell command to run (\texttt{\small locale -a}), the locales to prefer (\texttt{\small C.utf8}, \texttt{\small en\_US.utf8}),
and the exception to raise, and the complete hint for this task runs to thirteen times
the length of the Declaration one. 
These descriptions are produced by a language model that is shown the reference patch, so they are solution-derived by construction; the open question is how much of the solution survives into the text. 
We answer it with a language-model judge in Section~\ref{sec:judge}: the structural attributes carry no solution content, whereas the semantic ones describe behaviour and mechanism.
The ablation in Section~\ref{sec:ablation} shows that structural attributes deliver a resolve-rate gain at no disclosure, while the further gain from the semantic ones is bought with disclosed solution content. 
Declaration is therefore the largest projection we can justify as removing guesswork without supplying spoilers.

%% file: sections/experiments.tex
\section{Experiments}
\label{sec:experiments}
Our experiments examine how hint construction affects task resolution and what information it reveals about the reference solution.
We first isolate the effects of identifier selection and projection by varying one while holding the other fixed (RQ1 \& RQ2).
Then, we examine the attributes underlying these policies through ablations and an LLM-based assessment (RQ3 \& RQ4).
Together, these experiments assess the rationale for combining DefUse selection with Declaration projection.
We also repeat the selection and projection comparisons on Java to validate whether the patterns observed in Python extend to a second language.
\begin{itemize}[topsep=0mm,parsep=0mm]
    \item \textbf{RQ1} (selection, Section~\ref{sec:rq1}): For which identifiers should a
    hint expose information?
    \item \textbf{RQ2} (projection, Section~\ref{sec:rq2}): What information should a hint
    expose for each identifier?
    \item \textbf{RQ3} (ablation, Section~\ref{sec:ablation}): Which attributes of a hint carry its effect on task
    resolution?
    \item \textbf{RQ4} (judge, Section~\ref{sec:judge}): Which attributes of a hint reveal the reference solution?
\end{itemize}

\subsection{Implementation Details}
Our pipeline extracts, for every instance, the identifiers introduced or touched by the
reference patch and the test patch together with their attributes, and renders a hint from
a selection rule and a projection. To validate the extraction we reimplemented the
published hint pipelines of Doc2Feat-bench~\citep{deng2025nocode} and
FEA-Bench~\citep{li2025fea} and reproduced their released hints, with minor bug fixes. We also implemented our version of the SWE-Bench Pro hints pipeline by first extracting desired interfaces, and then using Claude Haiku 4.5~\citep{haiku45} to generate requirements and the interface descriptions.
Details available in Appendix~\ref{app:impl-details}.


\subsection{Settings}
\paragraph{Benchmarks, evaluation subsets, and selection statistics.}
We use four Python benchmarks: Doc2Feat-bench~\citep{deng2025nocode},
FEA-Bench~\citep{li2025fea}, SWE-bench Pro (denoted as Pro in tables) ~\citep{deng2025swe}, and
SWE-rebench~\citep{badertdinov_et_al_2025}. 
We additionally evaluate Java tasks from
Multi-SWE-bench (denoted as MSB Java)~\citep{zan2025multiswebench}.
For the selection comparisons (RQ1), we retain instances for which both the Doc2Feat and FEA-Bench Brief selection policies produce non-empty identifier sets.
All selection policies are evaluated on this shared subset, comprising $31$, $40$, $131$, and $26$ instances from the four Python benchmarks, respectively, and $40$ Java instances.
For the projection comparisons (RQ2), we retain instances for which DefUse produces a non-empty identifier set.
These subsets contain $21$, $44$, $68$, and $14$ instances from the four Python benchmarks, respectively, and $29$ Java instances.
The attribute ablations (RQ3) and LLM-based assessment (RQ4) focus on the same 68-instance SWE-bench Pro subset.
Table~\ref{tab:selection_counts} summarizes the number of identifiers disclosed by each selection policy.
DefUse discloses the fewest identifiers per instance across all five benchmarks, while Pro discloses the most.

\input{tables/selection_counts}


\paragraph{Models and agent.}
We run three language models through the same harness: DeepSeek V4 Flash~(0431), 
MiniMax M2.5, and GPT 5.6 Luna~\citep{deepseek,m25,gpt56luna}.
All models use temperature $1.0$, with maximum reasoning effort for DeepSeek and MiniMax and high reasoning effort for GPT 5.6 Luna.
We use mini-SWE-agent~\citep{yang2024sweagent}, which issues one shell command per turn, with a limit of 250 steps per trajectory.
Due to budget constraints, we run one trajectory per instance for each configuration.

\paragraph{Hint attributes.}
We parse each patch to extract the structural attributes, namely name, kind, path, new-file status and signature. 
The two semantic attributes mirror the human-written fields of
SWE-bench Pro: 
\emph{weak semantics} corresponds to the description under its
\texttt{interface}, 
and \emph{strong semantics} to its \texttt{requirements}. 
Both are generated by Claude Haiku 4.5~\citep{haiku45} given the reference patch, and both appear only in the Pro projection.

\paragraph{Evaluation and metrics.}
We evaluate each benchmark using its official evaluator in Docker containers. An instance
counts as resolved only if all of its fail-to-pass and pass-to-pass tests pass, and we
report the resolve rate, the percentage of resolved instances.

\subsection{Identifier Hint Selection}
\label{sec:rq1}
RQ1 examines which identifiers a hint should disclose.
We compare four selection policies under two fixed projections: 
Declaration (Table~\ref{tab:selection_declaration}),
which exposes structural interface attributes,
and Name Only (Appendix~\ref{app:selection-name-only},
Table~\ref{tab:selection}), which exposes only identifier names.
Because \ourselect{} restricts selection to identifiers through which the implementation and tests communicate (Section~\ref{sec:defuse}), 
these comparisons test whether
disclosing additional identifiers improves task resolution.

The selection policies differ substantially in how many identifiers they disclose (Table~\ref{tab:selection_counts}).
On SWE-bench Pro, \ourselect{} reveals an average of 1.2 identifiers per instance, compared with 4.8 for Doc2Feat, 7.7 for FEA, and 17.2 for Pro.
For GPT~5.6 Luna, this substantial difference in disclosure corresponds to a relatively small difference in resolution:
\ourselect{} resolves 91 of 131 instances~(69.5\%), compared with 92--95 (70.2--72.5\%) for the broader selections. Thus, for this model and benchmark, the narrower selection
substantially reduces disclosure with only a small decrease in observed resolution.

The performance cost of narrower disclosure is larger
for the open-source models on \mbox{SWE-bench Pro}.
DeepSeek resolves 17 instances with \ourselect{},
compared with 26--29 for the broader selection policies,
while MiniMax resolves 20, compared with 21--29.
These gains reach 12 and 9 additional instances,
respectively, compared with at most 4 for GPT~5.6 Luna.
However, the benefit of additional identifiers varies
across benchmarks.
On FEA-Bench, DeepSeek resolves 16 instances with
\ourselect{}, compared with 13--14 for the broader
policies, while MiniMax resolves 14, compared with 13--15.
On Multi-SWE-bench Java, DeepSeek resolves 13 of 40u
instances with \ourselect{} under Declaration,
compared with 12--15 for the other policies
(Table~\ref{tab:selection_java}).
Doc2Feat-bench remains challenging for both open-source
models, with DeepSeek resolving 0--2 instances and
MiniMax resolving one under every policy.
Together, these results indicate that the benefit of
broader disclosure depends on both the model and
the benchmark.

Broader selection can therefore improve measured resolution,
but its usefulness to an agent does not establish that
the additional identifiers are necessary to specify
the interface required by the tests.
Our motivation for \ourselect{} is to expose those
test-facing identifiers while limiting disclosure of
reference-specific implementation choices.

\input{tables/selection}

\input{tables/selection_java_ablation}

\subsection{Identifier Hint Projection}
\label{sec:rq2}
RQ2 examines what information a hint should disclose
about each selected identifier.
We hold selection fixed at \ourselect{} and compare
five projections against an empty baseline
(Table~\ref{tab:projection}).
This comparison examines how structural interface
attributes and additional semantic descriptions affect
task resolution.

The results show that structural interface information
can help agents resolve tasks.
On SWE-bench Pro, Declaration increases resolve rates
over the empty baseline from 10.3\% to 29.4\% for DeepSeek,
from 2.9\% to 14.7\% for MiniMax, and from 55.9\% to
73.5\% for GPT~5.6 Luna.
It also improves over Name Only for all three models
on this benchmark, suggesting that attributes beyond
identifier names contribute useful interface information.
These improvements occur without the semantic descriptions
included in the Pro projection.
The Java results provide supporting evidence:
Declaration increases DeepSeek's resolved instances
from 5 to 10 out of 29 (Table~\ref{tab:projection_java}).

The effect of adding semantic descriptions varies
across models.
On SWE-bench Pro, MiniMax resolves 16 instances with Pro,
compared with 10 under Declaration.
DeepSeek resolves 19 with Pro and 20 with Declaration,
while GPT~5.6 Luna resolves 39 and 50, respectively.
Thus, additional reference-derived descriptions can
improve resolution, but their benefit is not consistent
across models.
These comparisons alone do not establish whether the
descriptions help by clarifying requirements or by
supplying implementation guidance.
We also interpret small differences cautiously, given
the single trajectory per instance and the limited
sizes of some evaluation subsets.

Together, these observations support Declaration as
a reasonable projection for identifier hints:
it provides interface information that can improve
task resolution while omitting semantic descriptions
derived from the reference solution.
The attribute ablations and LLM-based assessment
(Sections~\ref{sec:ablation} and~\ref{sec:judge})
further examine the contributions of individual
attributes and the solution information they disclose.

\input{tables/projection_v2}
\input{tables/projection_java}
\subsection{Hint Attribute Ablation}
\label{sec:ablation}
RQ3 examines how different hint attributes contribute
to task resolution.
We fix selection to \ourselect{} and evaluate DeepSeek
V4 Flash on the SWE-bench Pro subset
(Table~\ref{tab:hint_attributes_ablation}).
The rows up to \emph{Declaration} successively add structural
attributes. \emph{Name Only and Semantics} combines names
with the two semantic descriptions, while \emph{All Hints}
combines Declaration with both descriptions.

Names improve resolution over the empty baseline,
and Declaration provides a further gain.
Adding entity type alone to names yields no observed
improvement, suggesting that structural attributes
differ in their contribution.
Semantic descriptions produce larger gains in these runs:
combining them with Declaration increases resolution
from 29.4\% to 72.1\%.
The combination also outperforms names with semantic
descriptions, suggesting that structural and semantic
information can provide complementary benefits.

These results distinguish the usefulness of hint
attributes from their appropriateness for evaluation.
Interface attributes can help agents satisfy the
test-facing contract, while semantic descriptions may
also supply implementation guidance.
Section~\ref{sec:judge} examines this distinction.
Each configuration uses a single trajectory per instance,
and the recorded variation across runs limits
attribution of the observed differences solely to hint
content.
The semantic configurations also use a different rendering
from Pro in RQ2, so their absolute rates should not be
directly compared with Table~\ref{tab:projection}.

\subsection{Hint Leakage Judge}
\label{sec:judge}
RQ4 examines what information hint attributes disclose,
complementing the task-resolution experiments.
Following prior work on LLM-based
evaluation~\citep{zheng2023judging,liu2023geval},
we use GPT~5.6 Luna with reasoning effort medium to assess attributes on SWE-bench Pro.
The rubric scores necessity $N$ (unrecoverability without
the reference solution), disclosure $D$ (solution content),
and guidance $G$ (direction of the agent's actions).
We summarize these ordinal scores using the heuristic hint value
margin $\mathrm{HVM}=N-\max(D,G)$.
The rubric and assessment procedure appear in
Appendix~\ref{app:hint-judge}.

Table~\ref{tab:hint_attribute_scores} shows similar
average necessity scores for structural and semantic
attributes, but greater disclosure and guidance for
semantic descriptions.
Names and signatures receive the highest margins,
whereas the semantic attributes have margins near zero.
Strong semantics receives higher disclosure and guidance
scores than weak semantics, consistent with descriptions
that convey implementation mechanisms.

Together with the projection comparison and attribute
ablation (Sections~\ref{sec:rq2} and~\ref{sec:ablation}),
this assessment supports Declaration as a reasonable
projection for identifier hints.
The experiments show that structural interface attributes
can improve task resolution without semantic descriptions,
while the judge associates those descriptions with greater
solution disclosure and implementation guidance.
Declaration therefore provides a principled boundary
for hint construction. 
It clarifies test-facing interfaces
while limiting additional reference-derived information
that may supply part of the implementation being evaluated.

\input{tables/judge_attributes}

%% file: tables/selection_counts.tex
\begin{table*}[t]
\centering
\caption{Number of identifiers disclosed per selection method, counted on the same instances as the selection tables under the Declaration projection. \emph{Total} is the number of identifiers over all instances and \emph{Mean} the number per instance. \ourselect{} is empty on the instances whose tests reference no new identifier: it is non-empty on 20/31, 44/142, 66/131, 9/26 and 29/40 instances, with 3.2, 1.8, 2.4, 2.0 and 6.7 identifiers per non-empty instance. The other selections are non-empty on every instance.}
\label{tab:selection_counts}

\small
\setlength{\tabcolsep}{4pt}
\begin{tabular}{lrrrrrrrrrr}
\toprule
\multirow{2}{*}{\textbf{Select}}
& \multicolumn{2}{c}{\textbf{Doc2Feat (31)}}
& \multicolumn{2}{c}{\textbf{FEA Bench (142)}}
& \multicolumn{2}{c}{\textbf{Pro (131)}}
& \multicolumn{2}{c}{\textbf{SWE Rebench (26)}}
& \multicolumn{2}{c}{\textbf{MSB Java (40)}} \\
\cmidrule(lr){2-3}
\cmidrule(lr){4-5}
\cmidrule(lr){6-7}
\cmidrule(lr){8-9}
\cmidrule(lr){10-11}
& \textcolor{gray}{\textbf{Total}} & \textbf{Mean}
  & \textcolor{gray}{\textbf{Total}} & \textbf{Mean}
  & \textcolor{gray}{\textbf{Total}} & \textbf{Mean}
  & \textcolor{gray}{\textbf{Total}} & \textbf{Mean}
  & \textcolor{gray}{\textbf{Total}} & \textbf{Mean} \\
\midrule
Empty    & \absval{0} & 0.0 & \absval{0} & 0.0 & \absval{0} & 0.0 & \absval{0} & 0.0 & \absval{0} & 0.0 \\
DefUse   & \absval{64} & 2.1 & \absval{79} & 0.6 & \absval{158} & 1.2 & \absval{18} & 0.7 & \absval{193} & 4.8 \\
Doc2Feat & \absval{173} & 5.6 & \absval{500} & 3.5 & \absval{635} & 4.8 & \absval{106} & 4.1 & \absval{313} & 7.8 \\
FEA      & \absval{337} & 10.9 & \absval{449} & 3.2 & \absval{1004} & 7.7 & \absval{110} & 4.2 & \absval{453} & 11.3 \\
Pro      & \absval{594} & 19.2 & \absval{1882} & 13.3 & \absval{2251} & 17.2 & \absval{418} & 16.1 & \absval{912} & 22.8 \\
\bottomrule
\end{tabular}
\end{table*}

%% file: tables/selection.tex
\begin{table*}[t]
\centering
\caption{Model performance across benchmarks and selection methods. We choose Declaration projection as the fixed projection given different identifiers.}
\label{tab:selection_declaration}

\small
\setlength{\tabcolsep}{4pt}
\centerline{\scriptsize\begin{tabular}{llcccccccc}
\toprule
\multirow{2}{*}{\textbf{Model}}
& \multirow{2}{*}{\textbf{Select}}
& \multicolumn{2}{c}{\textbf{Doc2Feat (31)}}
& \multicolumn{2}{c}{\textbf{FEA Bench (40)}}
& \multicolumn{2}{c}{\textbf{Pro (131)}}
& \multicolumn{2}{c}{\textbf{SWE Rebench (26)}} \\
\cmidrule(lr){3-4}
\cmidrule(lr){5-6}
\cmidrule(lr){7-8}
\cmidrule(lr){9-10}
& & \textcolor{gray}{\textbf{Resolved}} & \textbf{Acc.}
  & \textcolor{gray}{\textbf{Resolved}} & \textbf{Acc.}
  & \textcolor{gray}{\textbf{Resolved}} & \textbf{Acc.}
  & \textcolor{gray}{\textbf{Resolved}} & \textbf{Acc.} \\
\midrule

\multirow{5}{*}{DeepSeek V4 Flash}
& Empty    & \absval{0}  & 0.0\%  & \absval{13} & 32.5\% & \absval{11} & 8.4\%  & \absval{11} & 42.3\% \\
& DefUse   & \absval{1}  & 3.2\%  & \absval{16} & 40.0\% & \absval{17} & 13.0\% & \absval{9}  & 34.6\% \\
& Doc2Feat & \absval{0}  & 0.0\%  & \absval{14} & 35.0\% & \absval{26} & 19.8\% & \absval{2}  & 7.7\%  \\
& FEA      & \absval{2}  & 6.5\%  & \absval{13} & 32.5\% & \absval{27} & 20.6\% & \absval{6}  & 23.1\% \\
& Pro      & \absval{2}  & 6.5\%  & \absval{14} & 35.0\% & \absval{29} & 22.1\% & \absval{5}  & 19.2\% \\

\midrule

\multirow{5}{*}{MiniMax M2.5}
& Empty    & \absval{1}  & 3.2\%  & \absval{14} & 35.0\% & \absval{8}  & 6.1\%  & \absval{8}  & 30.8\% \\
& DefUse   & \absval{1}  & 3.2\%  & \absval{14} & 35.0\% & \absval{20} & 15.3\% & \absval{8}  & 30.8\% \\
& Doc2Feat & \absval{1}  & 3.2\%  & \absval{13} & 32.5\% & \absval{21} & 16.0\% & \absval{9}  & 34.6\% \\
& FEA      & \absval{1}  & 3.2\%  & \absval{15} & 37.5\% & \absval{27} & 20.6\% & \absval{13} & 50.0\% \\
& Pro      & \absval{1}  & 3.2\%  & \absval{15} & 37.5\% & \absval{29} & 22.1\% & \absval{13} & 50.0\% \\

\midrule

\multirow{5}{*}{GPT 5.6 Luna}
& Empty    & \absval{16} & 51.6\% & \absval{31} & 77.5\% & \absval{77} & 58.8\% & \absval{18} & 69.2\% \\
& DefUse   & \absval{17} & 54.8\% & \absval{30} & 75.0\% & \absval{91} & 69.5\% & \absval{18} & 69.2\% \\
& Doc2Feat & \absval{18} & 58.1\% & \absval{30} & 75.0\% & \absval{92} & 70.2\% & \absval{20} & 76.9\% \\
& FEA      & \absval{16} & 51.6\% & \absval{27} & 67.5\% & \absval{94} & 71.8\% & \absval{19} & 73.1\% \\
& Pro      & \absval{22} & 71.0\% & \absval{32} & 80.0\% & \absval{95} & 72.5\% & \absval{19} & 73.1\% \\

\bottomrule
\end{tabular}}
\end{table*}

%% file: tables/selection_java_ablation.tex
\begin{table*}[t]
\centering
\begin{minipage}[t]{0.48\textwidth}
\caption{DeepSeek V4 Flash performance on Multi-SWE-bench Java across selection policies under Declaration projection.}
\label{tab:selection_java}
\end{minipage}\hfill
\begin{minipage}[t]{0.48\textwidth}
\caption{Hint attribute ablation with DeepSeek V4 Flash on SWE-bench Pro (68 instances) under \ourselect{} selection.}
\label{tab:hint_attributes_ablation}
\end{minipage}
\par\smallskip
\begin{minipage}[t]{0.48\textwidth}
\centering
\scriptsize
\setlength{\tabcolsep}{4pt}
\begin{tabular}[t]{llcc}
\toprule
\multirow{2}{*}{\textbf{Project}}
& \multirow{2}{*}{\textbf{Select}}
& \multicolumn{2}{c}{\textbf{Multi-SWE-bench Java (40)}} \\
\cmidrule(lr){3-4}
& & \textcolor{gray}{\textbf{Resolved}} & \textbf{Acc.} \\
\midrule

\multirow{5}{*}{Declaration}
& Empty    & \absval{3} & 7.5\% \\
& DefUse   & \absval{13} & 32.5\% \\
& Doc2Feat & \absval{15} & 37.5\% \\
& FEA      & \absval{15} & 37.5\% \\
& Pro      & \absval{12} & 30.0\% \\
\bottomrule
\end{tabular}
\end{minipage}\hfill
\begin{minipage}[t]{0.48\textwidth}
\centering
\scriptsize
\setlength{\tabcolsep}{4pt}
\begin{tabular}[t]{lcc}
\toprule
\multirow{2}{*}{\textbf{Hint attributes}} & \multicolumn{2}{c}{\textbf{Pro (68)}} \\
\cmidrule(lr){2-3}
& \textcolor{gray}{\textbf{Resolved}} & \textbf{Acc.} \\
\midrule
Empty                   & \absval{7} & 10.3\% \\
Name Only               & \absval{15} & 22.1\% \\
Doc2Feat                & \absval{14} & 20.6\% \\
Declaration             & \absval{20} & 29.4\% \\
Name Only and Semantics & \absval{29} & 42.6\% \\
All Hints               & \absval{49} & 72.1\% \\
Oracle Hint             & \absval{60} & 88.2\% \\
\bottomrule
\end{tabular}
\end{minipage}
\end{table*}

%% file: tables/projection_v2.tex
\begin{table*}[t]
\centering
\caption{Model performance across benchmarks and projection methods. We choose \ourselect{} selection as the fixed selection given different projections. \emph{Declaration+Doc} adds a one-sentence description of each identifier to Declaration.}
\label{tab:projection}

\small
\setlength{\tabcolsep}{4pt}
\centerline{\scriptsize\begin{tabular}{llcccccccc}
\toprule
\multirow{2}{*}{\textbf{Model}}
& \multirow{2}{*}{\textbf{Project}}
& \multicolumn{2}{c}{\textbf{Doc2Feat (21)}}
& \multicolumn{2}{c}{\textbf{FEA Bench (44)}}
& \multicolumn{2}{c}{\textbf{Pro (68)}}
& \multicolumn{2}{c}{\textbf{SWE Rebench (14)}} \\
\cmidrule(lr){3-4}
\cmidrule(lr){5-6}
\cmidrule(lr){7-8}
\cmidrule(lr){9-10}
& & \textcolor{gray}{\textbf{Resolved}} & \textbf{Acc.}
  & \textcolor{gray}{\textbf{Resolved}} & \textbf{Acc.}
  & \textcolor{gray}{\textbf{Resolved}} & \textbf{Acc.}
  & \textcolor{gray}{\textbf{Resolved}} & \textbf{Acc.} \\
\midrule

\multirow{6}{*}{DeepSeek V4 Flash}
& Empty       & \absval{0} & 0.0\% & \absval{13} & 29.5\% & \absval{7} & 10.3\% & \absval{4} & 28.6\% \\
& NameOnly    & \absval{0} & 0.0\% & \absval{15} & 34.1\% & \absval{15} & 22.1\% & \absval{3} & 21.4\% \\
& Doc2Feat    & \absval{0} & 0.0\% & \absval{17} & 38.6\% & \absval{14} & 20.6\% & \absval{5} & 35.7\% \\
& Declaration & \absval{2} & 9.5\% & \absval{15} & 34.1\% & \absval{20} & 29.4\% & \absval{4} & 28.6\% \\
& Declaration+Doc & \absval{0} & 0.0\% & \absval{15} & 34.1\% & \absval{11} & 16.2\% & \absval{4} & 28.6\% \\
& Pro         & \absval{2} & 9.5\% & \absval{16} & 36.4\% & \absval{19} & 27.9\% & \absval{5} & 35.7\% \\

\midrule

\multirow{6}{*}{MiniMax M2.5}
& Empty       & \absval{1} & 4.8\% & \absval{15} & 34.1\% & \absval{2} & 2.9\% & \absval{4} & 28.6\% \\
& NameOnly    & \absval{0} & 0.0\% & \absval{18} & 40.9\% & \absval{7} & 10.3\% & \absval{5} & 35.7\% \\
& Doc2Feat    & \absval{0} & 0.0\% & \absval{14} & 31.8\% & \absval{8} & 11.8\% & \absval{4} & 28.6\% \\
& Declaration & \absval{2} & 9.5\% & \absval{15} & 34.1\% & \absval{10} & 14.7\% & \absval{4} & 28.6\% \\
& Declaration+Doc & \absval{0} & 0.0\% & \absval{14} & 31.8\% & \absval{11} & 16.2\% & \absval{3} & 21.4\% \\
& Pro         & \absval{0} & 0.0\% & \absval{17} & 38.6\% & \absval{16} & 23.5\% & \absval{4} & 28.6\% \\

\midrule

\multirow{6}{*}{GPT 5.6 Luna}
& Empty       & \absval{12} & 57.1\% & \absval{33} & 75.0\% & \absval{38} & 55.9\% & \absval{7} & 50.0\% \\
& NameOnly    & \absval{12} & 57.1\% & \absval{34} & 77.3\% & \absval{37} & 54.4\% & \absval{7} & 50.0\% \\
& Doc2Feat    & \absval{13} & 61.9\% & \absval{33} & 75.0\% & \absval{37} & 54.4\% & \absval{8} & 57.1\% \\
& Declaration & \absval{11} & 52.4\% & \absval{35} & 79.5\% & \absval{50} & 73.5\% & \absval{9} & 64.3\% \\
& Declaration+Doc & \absval{11} & 52.4\% & \absval{34} & 77.3\% & \absval{45} & 66.2\% & \absval{7} & 50.0\% \\
& Pro         & \absval{11} & 52.4\% & \absval{34} & 77.3\% & \absval{39} & 57.4\% & \absval{7} & 50.0\% \\

\bottomrule
\end{tabular}}
\end{table*}

%% file: tables/projection_java.tex
\begin{table*}[t]
\centering
\caption{Model performance on Multi-SWE-bench Java across projection methods. We choose \ourselect{} selection as the fixed selection given different projections. \emph{Declaration+Doc} adds a one-sentence description of each identifier to Declaration. Model: DeepSeek V4 Flash.}
\label{tab:projection_java}

\small
\setlength{\tabcolsep}{4pt}
\centerline{\scriptsize\begin{tabular}{llcc}
\toprule
\multirow{2}{*}{\textbf{Model}}
& \multirow{2}{*}{\textbf{Project}}
& \multicolumn{2}{c}{\textbf{Multi-SWE-bench Java (29)}} \\
\cmidrule(lr){3-4}
& & \textcolor{gray}{\textbf{Resolved}} & \textbf{Acc.} \\
\midrule

\multirow{6}{*}{DeepSeek V4 Flash}
& Empty       & \absval{5} & 17.2\% \\
& NameOnly    & \absval{10} & 34.5\% \\
& Doc2Feat    & \absval{7} & 24.1\% \\
& Declaration & \absval{10} & 34.5\% \\
& Declaration+Doc & \absval{9} & 31.0\% \\
& Pro         & \absval{14} & 48.3\% \\

\bottomrule
\end{tabular}}
\end{table*}

%% file: tables/judge_attributes.tex
\begin{table*}[t]
\centering
\caption{Hint attribute scores on SWE-bench Pro under
\ourselect{} selection. $N$: necessity; $D$: disclosure;
$G$: guidance; $\mathrm{HVM}=N-\max(D,G)$.
$^{*}$ marks margins indistinguishable from zero.
$\mathrm{HVM}$: Hint Value Margin.
Judge details appear in Appendix~\ref{app:hint-judge}.}
\label{tab:hint_attribute_scores}

\small
\setlength{\tabcolsep}{4pt}
\begin{tabular}{lrrrrr}
\toprule
\textbf{Attribute} & \textbf{n} & $N$ & $D$ & $G$ & \textbf{HVM} \\
\midrule
\multicolumn{6}{l}{\textit{Structural (interface-level)}} \\
\quad Signature & 31 & 2.45 & 0.00 & 0.00 & $+2.45$ \\
\quad Name & 160 & 1.88 & 0.00 & 0.11 & $+1.76$ \\
\quad Entity type & 160 & 1.54 & 0.00 & 0.00 & $+1.54$ \\
\quad Is new file & 58 & 0.62 & 0.00 & 0.00 & $+0.62$ \\
\quad Path & 58 & 1.19 & 0.00 & 1.00 & $+0.19$$^{*}$ \\
\quad \emph{group mean} & 467 & 1.56 & 0.00 & 0.16 & $\mathbf{+1.39}$ \\
\midrule
\multicolumn{6}{l}{\textit{Semantic (behaviour-level)}} \\
\quad Weak semantics & 160 & 1.19 & 1.09 & 0.29 & $+0.03$$^{*}$ \\
\quad Strong semantics & 160 & 1.82 & 1.81 & 1.68 & $-0.13$$^{*}$ \\
\quad \emph{group mean} & 320 & 1.50 & 1.45 & 0.99 & $\mathbf{-0.05}$ \\
\bottomrule
\end{tabular}
\end{table*}

%% file: sections/conclusions.tex
\section{Conclusions and Limitations}
\label{sec:conclusions}
We study identifier hints through selection and projection,
showing that hint policies affect both measured task
resolution and the reference-derived information available
to agents.
Our findings motivate combining \ourselect{} selection
with Declaration projection to clarify test-facing
interfaces while limiting implementation guidance.
More broadly, benchmark designers should assess hints
by what they disclose as well as the performance gains
they produce.

\paragraph{Limitations.}
Budget constraints limited each reported configuration
to one trajectory per instance, making small differences
sensitive to stochastic variation.
The recorded experiments represent approximately \$2,300
in inference value, including unbilled MiniMax usage
valued at public API rates.
Small benchmark subsets and a single agent harness limit
generalizability, while serving variation and rubric-dependent
LLM judgments introduce additional uncertainty.

%% file: sections/appendix.tex
\input{sections/appendix/implementation_details}

\input{sections/appendix/selection_name_only}
\input{sections/appendix/judge}

%% file: sections/appendix/implementation_details.tex
\section{Implementation Details: Baselines}
\label{app:impl-details}

\subsection{Selection}
For both Doc2Feat and FEA Bench Brief hint generation, we adapted the original code so that the pipeline can extract hints on various benchmarks.
For SWE Bench Pro hint generation, since they employ human engineers to build the hints, it's impossible to completely replicate Pro's pipeline. 
Thus, we use a LLM to approximate human engineers in terms of providing detailed hint text.
We first find the different identifiers between old code and new cold patches, extract the information about identifiers, and then have Claude Haiku 4.5 to generate interface descriptions and requirements with few shot prompts.
Comparing the generated interface descriptions and requirements with original ones in python subset of SWE Bench Pro (266 instances), the generated ones achieved consine similarity of $0.8978$ and $0.8405$ respectively, using Qwen3-0.6-Embedding~\citep{qwen3embedding} as the embedding model.

\subsection{Projection}
To implement controllable projection, we first have a complete pipeline that extracts all possible identifiers from both the code patches and the test patches. 
For each identifier, we extract relevant information by checking its abstract syntax tree to get the information not just about the identifier name.
Then, we select DefUse as our selection by performing the communication between code and test patches, and the we adopt Claude Haiku 4.5 to enrich the semantics, where weak semantics corresponds to SWE Bench Pro interface descriptions, and strong semantics corresponds to requirements.
Then:
\begin{itemize}
    \item Doc2Feat: the name and entity type of each identifier, rendered as a flat
    list; no path, signature or description.
    \item Declaration: name, entity type, file path, a new-file flag and the
    signature; class members are nested under their class. No natural-language text.
    \item Declaration+Doc: Declaration plus the weak semantics, a one-sentence
    (at most 20 words) description of what the entity is and does. It corresponds to
    the docstring that FEA-Bench's detailed hints add, without their code; FEA-Bench's
    brief hints carry no description and coincide with Declaration.
    \item Pro: Declaration+Doc plus the strong semantics, a two-to-four sentence
    paragraph on behaviour, edge cases and test-implied constraints. Both descriptions
    are written by Claude Haiku 4.5 from the reference patch, so the LLM acts as the
    extractor that SWE-bench Pro's human annotators were.
\end{itemize}

%% file: sections/appendix/selection_name_only.tex
\section{Identifier Selection under Name Only Projection}
\label{app:selection-name-only}
We repeat the identifier selection comparison with the projection fixed at Name Only.
Each hint reveals only the names of the selected identifiers, allowing us to examine
selection policies without disclosing additional attributes.
Tables~\ref{tab:selection} and~\ref{tab:selection_java_nameonly} report the Python
and Java results, respectively.

The effect of selection varies across models and benchmarks even when hints contain
only names. On SWE-bench Pro, Pro and FEA outperform \ourselect{} for all three models,
whereas \ourselect{} leads the hint selections for DeepSeek on FEA-Bench.
The Java results provide a further example: \ourselect{} and Pro each resolve
9 of 40 instances, compared with 6 for Doc2Feat and FEA.
Thus, broader selection can help, but disclosing additional names does not consistently
improve resolution. As in the Declaration comparison, small differences should be
interpreted cautiously given the limited evaluation subsets and single trajectory
per instance.

\input{tables/selection_nameonly}
\input{tables/selection_java_nameonly}


%% file: tables/selection_nameonly.tex
\begin{table*}[ht]
\centering
\caption{Model performance across benchmarks and selection methods. We choose NameOnly Bench projection as the fixed projection given different identifiers.}
\label{tab:selection}

\small
\setlength{\tabcolsep}{4pt}
\centerline{\scriptsize\begin{tabular}{llcccccccc}
\toprule
\multirow{2}{*}{\textbf{Model}}
& \multirow{2}{*}{\textbf{Select}}
& \multicolumn{2}{c}{\textbf{Doc2Feat (31)}}
& \multicolumn{2}{c}{\textbf{FEA Bench (40)}}
& \multicolumn{2}{c}{\textbf{Pro (131)}}
& \multicolumn{2}{c}{\textbf{SWE Rebench (26)}} \\
\cmidrule(lr){3-4}
\cmidrule(lr){5-6}
\cmidrule(lr){7-8}
\cmidrule(lr){9-10}
& & \textcolor{gray}{\textbf{Resolved}} & \textbf{Acc.}
  & \textcolor{gray}{\textbf{Resolved}} & \textbf{Acc.}
  & \textcolor{gray}{\textbf{Resolved}} & \textbf{Acc.}
  & \textcolor{gray}{\textbf{Resolved}} & \textbf{Acc.} \\
\midrule

\multirow{5}{*}{DeepSeek V4 Flash}
& Empty    & \absval{0} & 0.0\%  & \absval{13} & 32.5\% & \absval{14} & 10.7\% & \absval{12} & 46.2\% \\
& DefUse   & \absval{2} & 6.7\%  & \absval{16} & 40.0\% & \absval{17} & 13.0\% & \absval{11} & 42.3\% \\
& Doc2Feat & \absval{1} & 3.3\%  & \absval{14} & 35.0\% & \absval{17} & 13.0\% & \absval{11} & 42.3\% \\
& FEA      & \absval{0} & 0.0\%  & \absval{13} & 32.5\% & \absval{20} & 15.3\% & \absval{11} & 42.3\% \\
& Pro      & \absval{1} & 3.3\%  & \absval{14} & 35.0\% & \absval{20} & 15.3\% & \absval{12} & 46.2\% \\

\midrule

\multirow{5}{*}{MiniMax M2.5}
& Empty    & \absval{1} & 3.3\%  & \absval{14} & 35.0\% & \absval{12} & 9.2\%  & \absval{8}  & 30.8\% \\
& DefUse   & \absval{2} & 6.7\%  & \absval{14} & 35.0\% & \absval{15} & 11.5\% & \absval{8}  & 30.8\% \\
& Doc2Feat & \absval{1} & 3.3\%  & \absval{13} & 32.5\% & \absval{16} & 12.2\% & \absval{10} & 38.5\% \\
& FEA      & \absval{0} & 0.0\%  & \absval{15} & 37.5\% & \absval{20} & 15.3\% & \absval{13} & 50.0\% \\
& Pro      & \absval{0} & 0.0\%  & \absval{15} & 37.5\% & \absval{20} & 15.3\% & \absval{10} & 38.5\% \\

\midrule

\multirow{5}{*}{GPT 5.6 Luna}
& Empty    & \absval{21} & 66.7\% & \absval{31} & 77.5\% & \absval{85} & 64.9\% & \absval{19} & 73.1\% \\
& DefUse   & \absval{22} & 70.0\% & \absval{30} & 75.0\% & \absval{83} & 63.4\% & \absval{19} & 73.1\% \\
& Doc2Feat & \absval{21} & 66.7\% & \absval{30} & 75.0\% & \absval{88} & 67.2\% & \absval{19} & 73.1\% \\
& FEA      & \absval{17} & 53.3\% & \absval{27} & 67.5\% & \absval{88} & 67.2\% & \absval{20} & 76.9\% \\
& Pro      & \absval{16} & 50.0\% & \absval{32} & 80.0\% & \absval{93} & 71.0\% & \absval{18} & 69.2\% \\

\bottomrule
\end{tabular}}
\end{table*}

%% file: tables/selection_java_nameonly.tex
\begin{table*}[t]
\centering
\caption{DeepSeek V4 Flash performance on Multi-SWE-bench Java across selection policies under Name Only projection.}
\label{tab:selection_java_nameonly}

\small
\setlength{\tabcolsep}{4pt}
\centerline{\scriptsize
\begin{tabular}{llcc}
\toprule
\multirow{2}{*}{\textbf{Project}}
& \multirow{2}{*}{\textbf{Select}}
& \multicolumn{2}{c}{\textbf{Multi-SWE-bench Java (40)}} \\
\cmidrule(lr){3-4}
& & \textcolor{gray}{\textbf{Resolved}} & \textbf{Acc.} \\
\midrule

\multirow{5}{*}{Name Only}
& Empty    & \absval{3} & 7.5\% \\
& DefUse   & \absval{9} & 22.5\% \\
& Doc2Feat & \absval{6} & 15.0\% \\
& FEA      & \absval{6} & 15.0\% \\
& Pro      & \absval{9} & 22.5\% \\
\bottomrule
\end{tabular}}
\end{table*}

%% file: sections/appendix/judge.tex
\section{Hint Leakage Judge Details}
\label{app:hint-judge}

\paragraph{Rubric.}
Each attribute is assessed on three ordinal scales.
Necessity $N$ measures how recoverable its value is without
the reference solution: 0 if stated in the issue,
1 if findable in the repository, 2 if constrained by
convention, and 3 if freely chosen by the reference author.
Disclosure $D$ measures solution content: 0 for identity
and addressing, 1 for the observable contract,
2 for an implementation mechanism, and 3 for a transcription.
Guidance $G$ measures direction of the agent's actions:
0 for declarative information, 1 for orientation,
2 for procedural guidance, and 3 for prescriptive instructions.
Disclosure and guidance are distinct: a file path can
orient the agent without describing an implementation.
The heuristic margin $\mathrm{HVM}=N-\max(D,G)$ combines
these judgments; it is not a calibrated measure of
information quantity.

\paragraph{Assessment procedure.}
We assess 787 identifier--attribute pairs across
160 identifiers selected by \ourselect{} on
68 SWE-bench Pro instances.
Necessity is assessed using the issue without showing
the reference or test patches; disclosure and guidance
are assessed with both patches available.
Each call presents all attributes of an identifier and
specifies one target attribute, so the necessity judgment
is conditional on the other visible attributes.
The judge is GPT~5.6 Luna with medium reasoning effort.
It returns a category label, an evidence span, and
a rationale.
We map labels to integers and take the median of three
samples at temperature~1.
We estimate uncertainty using instance-clustered
bootstrap and measure consistency using three-sample
unanimity.

\paragraph{Validation and limitations.}
We check repository membership for a subset of identifier
names using \texttt{git grep} at the base commit.
When supplied with this evidence, the judge reproduces
the repository-membership distinction for all 31
identifiers checked, compared with 18 without it.
This check validates a limited aspect of the necessity
assessment; it does not validate the full rubric.
Agreement also varies across attributes:
three-sample unanimity is 94\% for signatures but
41--42\% for semantic descriptions.
These differences indicate greater judgment uncertainty
for semantic content and should be considered alongside
the aggregate scores.

\paragraph{Rubrics and prompts.}
The three blocks below reproduce, verbatim from the judge implementation, the rubric
and the prompt skeleton behind each column of Table~\ref{tab:hint_attribute_scores}.
Necessity is one call per (instance, identifier, attribute) triple that sees the issue
but neither patch; disclosure and guidance are rated together in a second call that
also sees the reference and test patches. Angle-bracketed fields are filled per call.
The judge returns category labels only; the integers in the table are assigned in
code (\texttt{REDUNDANT}\,$\to$\,0, \ldots, \texttt{LOAD-BEARING}\,$\to$\,3, and
likewise for the other two scales). All three calls share the system prompt shown
at the top of the first block.

\begin{promptbox}{Necessity ($N$): rubric and solution-blind prompt}
[system]
You are an exacting research assistant evaluating hint quality for a software
engineering benchmark. You apply a rubric literally and consistently. You do not
reward values for being detailed or useful; you classify them by the stated
criteria only. You always answer with a single JSON object and nothing else.

[user]
Rate NECESSITY: could a competent engineer obtain this attribute's value WITHOUT
seeing the reference solution?

  0 REDUNDANT      - the value is stated in the problem statement itself.
  1 SEARCH-SAVING  - not stated, but an engineer with the repository at its current
                     commit would reach it through normal investigation (the entity
                     already exists; grep/reading finds it). The hint saves effort,
                     not information.
  2 DISAMBIGUATING - not stated and not already present, but strongly constrained by
                     repository convention (e.g. a new test file's location follows
                     an obvious existing layout). An engineer would likely guess it.
  3 LOAD-BEARING   - encodes a free choice the solution's author made that is not
                     recoverable from the problem statement or the existing codebase
                     (e.g. the name of a function that does not yet exist). Without
                     the hint an engineer would pick something else.

Judge ONLY the target attribute's value. Ignore whether the value seems important;
necessity is about RECOVERABILITY, not importance.

## Measured repository fact            (present only when the target is `name`)
The identifier `<name>` appears in <k> file(s) in this repository at the commit
the agent starts from.
<k > 0: "An engineer could therefore find it by searching the existing code."
 k = 0: "It does not exist anywhere in the tree yet, so it cannot be found by
         searching; any engineer would have to choose a name.">

## Repository
<owner/repo>

## Problem statement (this is ALL the agent is told)
<issue text, first 6000 characters>

## The identifier, for context
  name: <...>
  entity_type: <...>
  path: <...>                 <<< TARGET      (one attribute is marked)
  is_new_file: <...>
  signature: <...>
  weak_semantics: <...>
  strong_semantics: <...>

## The value you are rating
attribute : `<target attribute>`
value     : ```
<the target attribute's value, alone>
```

You are rating THE TEXT ABOVE - not the entity it refers to, and not the other
attributes listed for context. Ask what this text, by itself, tells a reader.

## Your task
Rate NECESSITY.

Respond with exactly this JSON object:
{"label": "<REDUNDANT|SEARCH-SAVING|DISAMBIGUATING|LOAD-BEARING>",
  "evidence": "<a short verbatim quote from the problem statement that justifies
your rating, or the empty string if the statement says nothing relevant>",
  "rationale": "<one sentence>",
  "confidence": <0.0-1.0>}
\end{promptbox}

\begin{promptbox}{Disclosure ($D$): rubric}
Rate DISCLOSURE: how much of the reference solution's CONTENT does this value carry?

  0 INTERFACE      - identity and addressing only: what it is called, where it lives,
                     how it is invoked. This is the line the benchmark itself
                     publishes to agents, so it is not leakage.
  1 CONTRACT       - observable behaviour: what it does, what its inputs and outputs
                     mean. A specification someone could implement many ways.
  2 MECHANISM      - how it does it: control flow, algorithm, ordering of operations,
                     or the names of specific helpers/APIs it calls internally.
  3 TRANSCRIPTION  - restates the solution closely enough that the code could be
                     transliterated from the text with little independent thought.

Example of 2: "overrides add_job to check the hostname against allowed hosts using
fnmatch patterns" - names the method and the matching library.
Example of 1: "A set of exact key names that should not be sanitized even if they
contain sensitive substrings" - states the contract without the mechanism.
\end{promptbox}

\begin{promptbox}{Guidance ($G$): rubric, and the solution-aware prompt shared with $D$}
Rate BEHAVIOURAL GUIDANCE: how much does this value DIRECT the agent's actions, as
opposed to merely stating a fact it must act upon?

  0 DECLARATIVE    - states that something exists or what it is called. No action
                     is implied.
  1 ORIENTING      - narrows where to look or what to touch, without saying what to do.
  2 PROCEDURAL     - implies a sequence of steps or a specific approach to take.
  3 PRESCRIPTIVE   - reads as an instruction the agent can follow directly; the text
                     could substitute for a plan.

This is INDEPENDENT of disclosure. A line range discloses a lot (exactly where the
change goes) but guides nothing (it does not say what to do). An algorithm
description does both. Score them separately.

------------------------------------------------------------------------------
[user message of the D/G call: the DISCLOSURE rubric, then the GUIDANCE rubric,
 then the following]

## Problem statement
<issue text, first 4000 characters>

## Reference solution (gold patch)
```diff
<gold patch, first 14000 characters>
```

## Test patch (defines what must pass)
```diff
<test patch, first 6000 characters>
```

## The identifier, for context
<all attributes of the identifier, target marked "<<< TARGET", as above>

## The value you are rating
<the target attribute's value, alone, as above>

## Your task
Rate DISCLOSURE and BEHAVIOURAL GUIDANCE for the value above, by comparing it
against the reference solution.

Respond with exactly this JSON object:
{"disclosure_label": "<INTERFACE|CONTRACT|MECHANISM|TRANSCRIPTION>",
  "guidance_label": "<DECLARATIVE|ORIENTING|PROCEDURAL|PRESCRIPTIVE>",
  "evidence": "<the span of the target value that carries the most solution content,
verbatim>",
  "rationale": "<one sentence>",
  "confidence": <0.0-1.0>}
\end{promptbox}